%% file: main.tex
\documentclass[11pt,a4paper]{article}
\usepackage[hyperref]{emnlp2018}
\usepackage{hyperref}
\usepackage{url}
\usepackage{times}
\usepackage{latexsym}
\usepackage{array}
\usepackage{multirow}
\usepackage{amssymb}
\usepackage{amsmath}
\usepackage{mathpartir}
\usepackage{mathtools}
\usepackage{amsthm}
\usepackage{xspace}
\usepackage{algorithm}
\usepackage{tikz-dependency}
\usepackage{cancel}
\usepackage{wrapfig}
\usepackage{dsfont}
\usepackage{booktabs}

\input{defs}
\aclfinalcopy 

\title{Breaking the Beam Search Curse: A Study of (Re-)Scoring Methods and Stopping Criteria for Neural Machine Translation}

\author{{ \hspace{7cm} Yilin Yang$^1$ \qquad\; Liang Huang$^{1,2}$ \qquad\; Mingbo Ma$^{1,2}$ }\\
  $^1$
  Oregon State University\\
  Corvallis, OR, USA\\
  {\hspace{7cm} \tt \small \{yilinyang721, liang.huang.sh, cosmmb\}@gmail.com} \And
  \\
  $^2$
  Baidu Research\\
  Sunnyvale, CA, USA\\
}
\date{}

\begin{document}
\maketitle
\begin{abstract}
Beam search is widely used in neural machine translation,
and usually improves translation quality compared to greedy search.
It has been widely observed that, however, beam sizes larger than 5 hurt translation quality. 
We explain why this happens,
and propose several methods to address this problem.
Furthermore, we discuss the optimal stopping criteria for these methods.
Results show that our hyperparameter-free methods outperform
the widely-used hyperparameter-free heuristic of length normalization by +2.0 BLEU, and achieve the best results among all methods on Chinese-to-English translation.
\end{abstract}

\section{Introduction}
\label{sec:intro}
\input{intro}

\section{Preliminaries: NMT and Beam Search}
\label{sec:background}
\input{background}

\section{Beam Search Curse}
\label{sec:curse}
\input{curse}

\section{Rescoring Methods}
\label{sec:methods}
\input{methods}

\section{Stopping Criteria}
\label{sec:stop}
\input{stop}

\section{Experiments}
\label{sec:exps}
\input{exps}

\section{Conclusions}
We first explain why the beam search curse exists
and then formalize all previous rescoring methods.
Beyond that, we also propose several new methods to address this problem.
Results from the Chinese-English task show that our hyperparameter-free methods beat the
hyperparameter-free baseline (length normalization) by +2.0 BLEU.

\section*{Acknowledgements}
Kenton Lee suggested the length prediction idea. 
This work was partially supported by DARPA N66001-17-2-4030, and NSF IIS-1817231 and IIS-1656051.
We thank the anonymous reviewers for suggestions and Juneki Hong for proofreading.

\bibliography{emnlp2018}
\bibliographystyle{acl_natbib_nourl}

\end{document}

%% file: defs.tex
\newcommand{\tuple}[1]{\ensuremath{\langle {#1} \rangle}}

\newcommand{\notes}[1]{}

\theoremstyle{definition}

\theoremstyle{plain}

\newcommand{\ith}[1]{\ensuremath{i^{{th}}}}

\newcount\permx
\newcount\permy
\def\permdot#1#2{
\permx=#1 \advance\permx by-1
\permy=#2 \advance\permy by-1
\psframe[fillcolor=black, fillstyle=solid]
(\permx,\permy)(#1, #2)
}

\newcommand{\toptop}{\operatornamewithlimits{\mathbf{top}}}

\newcommand{\startsym}{\mbox{\scriptsize \texttt{<s>}}\xspace}

\newcommand{\boxnum}[1]{{\setlength{\fboxsep}{1pt}\raisebox{1pt}{\hspace{1pt}\fbox{\tiny #1}\hspace{1pt}}}}
\newcommand{\ind}[1]{\ensuremath{_{\kern-0.5pt\boxnum{#1}}}}

\newcommand{\vecx}{\ensuremath{\mathbf{x}}\xspace}
\newcommand{\vecy}{\ensuremath{\mathbf{y}}\xspace}

\def\namecite{\newcite}

\newcommand{\smallnt}[1]{\ensuremath{_{\mbox{\tiny PP}}}\xspace}

\newcommand{\pseudocode}{Algorithm}
\floatname{algorithm}{\pseudocode}

\iffalse

\else

\fi

\newcommand{\score}{\ensuremath{\mathit{S}}\xspace}

\newcommand{\eos}{\mbox{\scriptsize \texttt{</eos>}}\xspace}
\newcommand{\bp}{\ensuremath{\mathit{bp}}\xspace}
\newcommand{\BLEU}{\ensuremath{\mathrm{BLEU}}\xspace}

\newcommand{\AdaR}{\ensuremath{\mathrm{AdaR}}\xspace}

\newcommand{\lengthnorm}{\ensuremath{\mathrm{length\_norm}}\xspace}
\newcommand{\word}{\ensuremath{\mathrm{word}}\xspace}

\newcommand{\gratio}{\ensuremath{\mathit{gr}}\xspace}
\newcommand{\lratio}{\ensuremath{\mathit{lr}}\xspace}
\newcommand{\pred}{\ensuremath{\mathit{pred}}\xspace}

%% file: intro.tex

In recent years, neural machine translation (NMT) has surpassed traditional
phrase-based or syntax-based machine translation, becoming the new
state of the art in MT
\cite{kalchbrenner+blunsom:2013,sutskever+:2014,bahdanau+:2014}.
While NMT training is typically done in a ``local'' fashion which does not employ any search (bar notable exceptions such as \namecite{ranzato+:2016}, \namecite{shen+:2016}, and \namecite{wiseman+rush:2016}),
the decoding phase of all NMT systems
universally adopts beam search, a widely used heuristic, to improve translation
quality. 

Unlike phrase-based MT systems which enjoy the benefits of very large beam sizes
(in the order of 100--500)~\cite{koehn+:2007} , 
most NMT systems choose tiny beam sizes up to 5;
for example, Google's GNMT~\cite{wu+:2016}
and Facebook's ConvS2S~\cite{gehring+:2017} use beam sizes~3 and 5, respectively.
Intuitively, the larger the beam size is, the more candidates it explores,
and the better the translation quality should be.
While this definitely holds for phrase-based MT systems,
surprisingly, it is {\em not} the case for NMT:
many researchers observe that translation quality
degrades with beam sizes beyond 5 or 10~\cite{tu+:2016,koehn+knowles:2017}.
We call this phenomenon the {\em ``beam search curse''},
which is listed as one of the six biggest challenges for NMT~\cite{koehn+knowles:2017}.

However, there has not been enough attention on this problem. 
\namecite{huang+:2017} hint that length ratio is the problem,
but do not explain why larger beam sizes cause shorter lengths and worse BLEU.
\namecite{Ott+:2018} attribute it to two kinds of ``uncertainties'' in the training data,
namely the {\em copying} of source sentence and the {\em non-literal} translations.
However, the first problem is only found in European language datasets
and the second problem occurs in all datasets but does not seem to bother pre-neural MT systems. 
Therefore, their explanations are not satisfactory.

On the other hand, previous work adopts several heuristics to address this problem,
but with various limitations.
For example, RNNSearch~\cite{bahdanau+:2014} and ConvS2S use {\em length normalization},
which (we will show in Sec.~\ref{sec:exps}) seems to somewhat alleviate the problem, but far from being perfect.
Meanwhile, \namecite{he+:2016} and \namecite{huang+:2017} use word-reward,
but their \textit{reward} is a hyper-parameter to be tuned on dev set. 

Our contributions are as follows:
\begin{itemize}
\item We explain why the beam search curse exists,
supported by empirical evidence (Sec.~\ref{sec:curse}).
\item We review existing rescoring methods,
and then propose ours to break the beam search curse (Sec.~\ref{sec:methods}).
We show that our hyperparameter-free methods outperfrom the previous
hyperparameter-free method (length normalization) by +2.0 BLEU (Sec.~\ref{sec:exps}).
\item We also discuss the stopping criteria for our rescoring methods (Sec.~\ref{sec:stop}).
Experiments show that with optimal stopping alone, the translation quality of
the length normalization method improves by +0.9 BLEU.
\end{itemize}

After we finish our paper, we became aware of a parallel work \cite{murray+chiang:2018}
that also reveals the same root cause we found for the beam search curse: the length ratio problem.


%% file: background.tex

We briefly review the encoder-decoder architecture with attention mechanism ~\cite{bahdanau+:2014}.
An RNN encoder takes an input sequence  $\vecx = (x_1,...,x_m)$,
and produces a sequence of hidden states.
For each time step, the RNN decoder will predict the probability of next output word
given the source sequence and the previously generated prefix.
Therefore, when doing greedy search, at time step $i$,
the decoder will choose the word with highest
probability as $y_i$.
The decoder will continue generating until it emits \eos .
In the end, the generated hypothesis is $\vecy = (y_1,...,y_n)$ with $y_n\!=\!\eos$, with model score
\begin{equation}
\vspace{-0.1cm}
\score(\vecx,\vecy) = \textstyle\sum_{i=1}^{|\vecy|} \log p(y_i \mid \vecx,\, y_{1..\{i-1\}})
\end{equation}


As greedy search 
only explores a single path,
we always use beam search to improve search quality.
Let $b$ denote the beam size, then at step $i$ the beam $B_i$
is an {\it ordered} list of size $b$: 
\begin{align*}
B_0\! &= \![\tuple{\startsym, \ p(\startsym \mid \vecx)}] \\
B_i\! &= \!\toptop^b
     \{\tuple{\vecy'\!\!\circ y_i, \ s\!\cdot\! p(y_i | \vecx, \vecy)} \mid \tuple{\vecy'\!, s} \in B_{i-1} \}
\end{align*}
In the most naive case, after reaching the maximum length (a hard limit), we get $N$ possible candidate sequences $\{\vecy_1,...,\vecy_N\}$.
The \textit{default} strategy chooses the one with highest model score.
We will discuss more sophistcated ways of stopping and choosing candidates in later sections.

%% file: curse.tex

\begin{figure}[t]\centering
\includegraphics[width=0.48\textwidth]{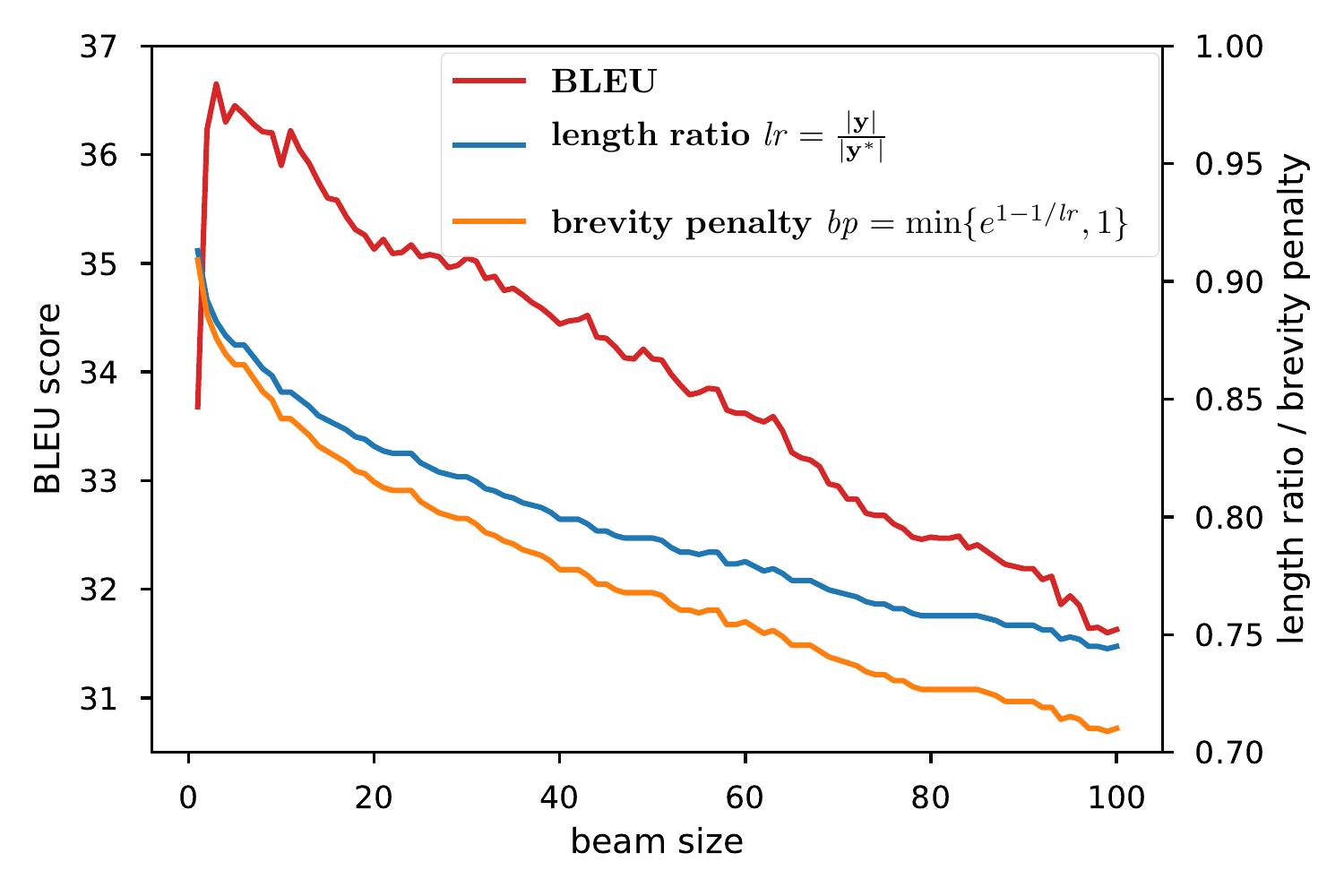}
\vspace{-0.9cm}
\caption{As beam size increases beyond 3, BLEU score on the dev set gradually drops.
  All terms are calculated by \textit{multi-bleu.pl}.}
\label{fig:curse-bleu}
\end{figure}

\begin{figure}[t]\centering
  \vspace{-0.3cm}
\includegraphics[width=0.45\textwidth]{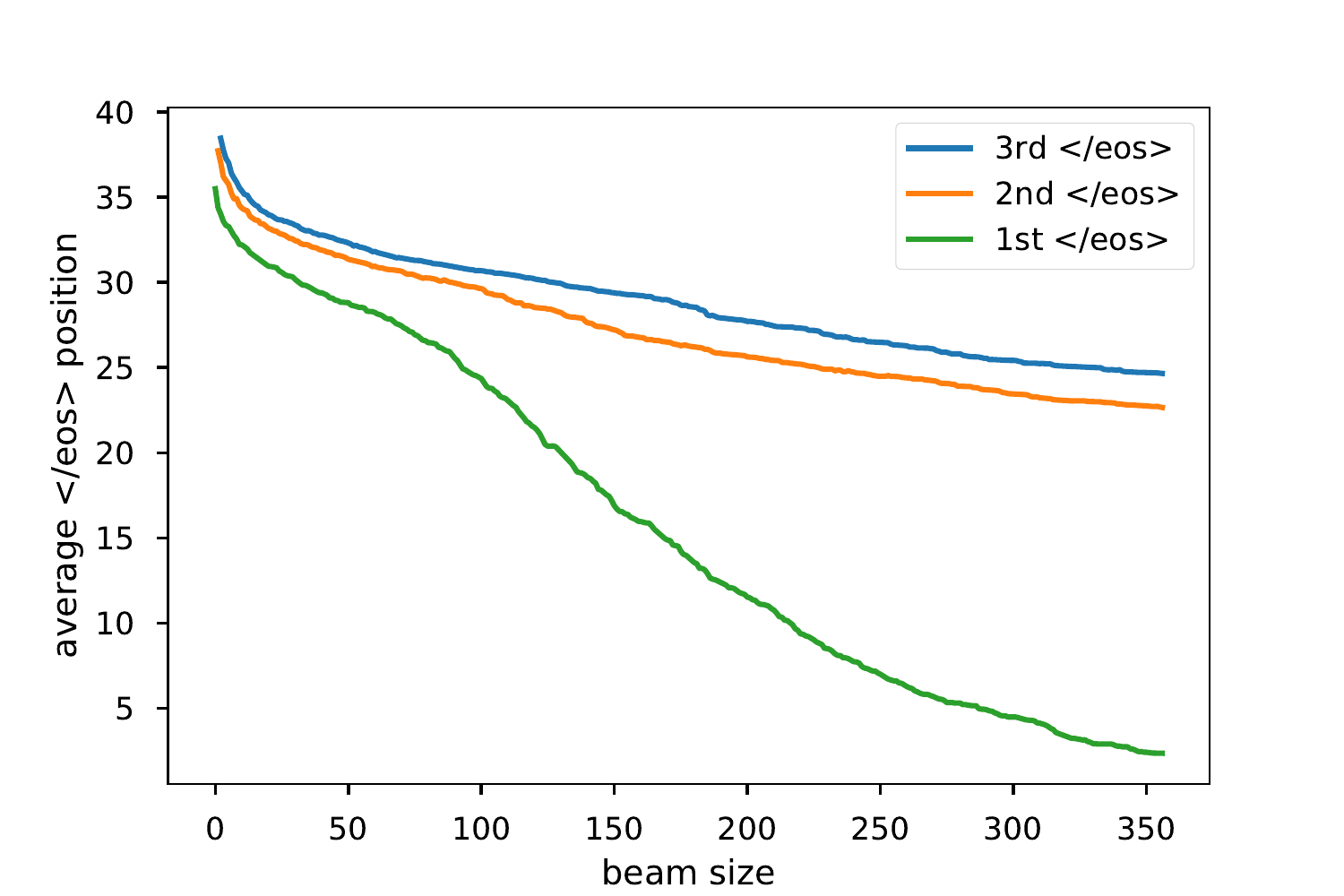}
\vspace{-0.3cm}
\caption{Searching algorithm with larger beams generates \eos earlier.
  We use the average first, second and third \eos positions on the dev set as an example.}
\label{fig:average_length-beam}
\end{figure}

\begin{figure}[t]\centering
  \vspace{-0.35cm}
  \includegraphics[width=0.48\textwidth]{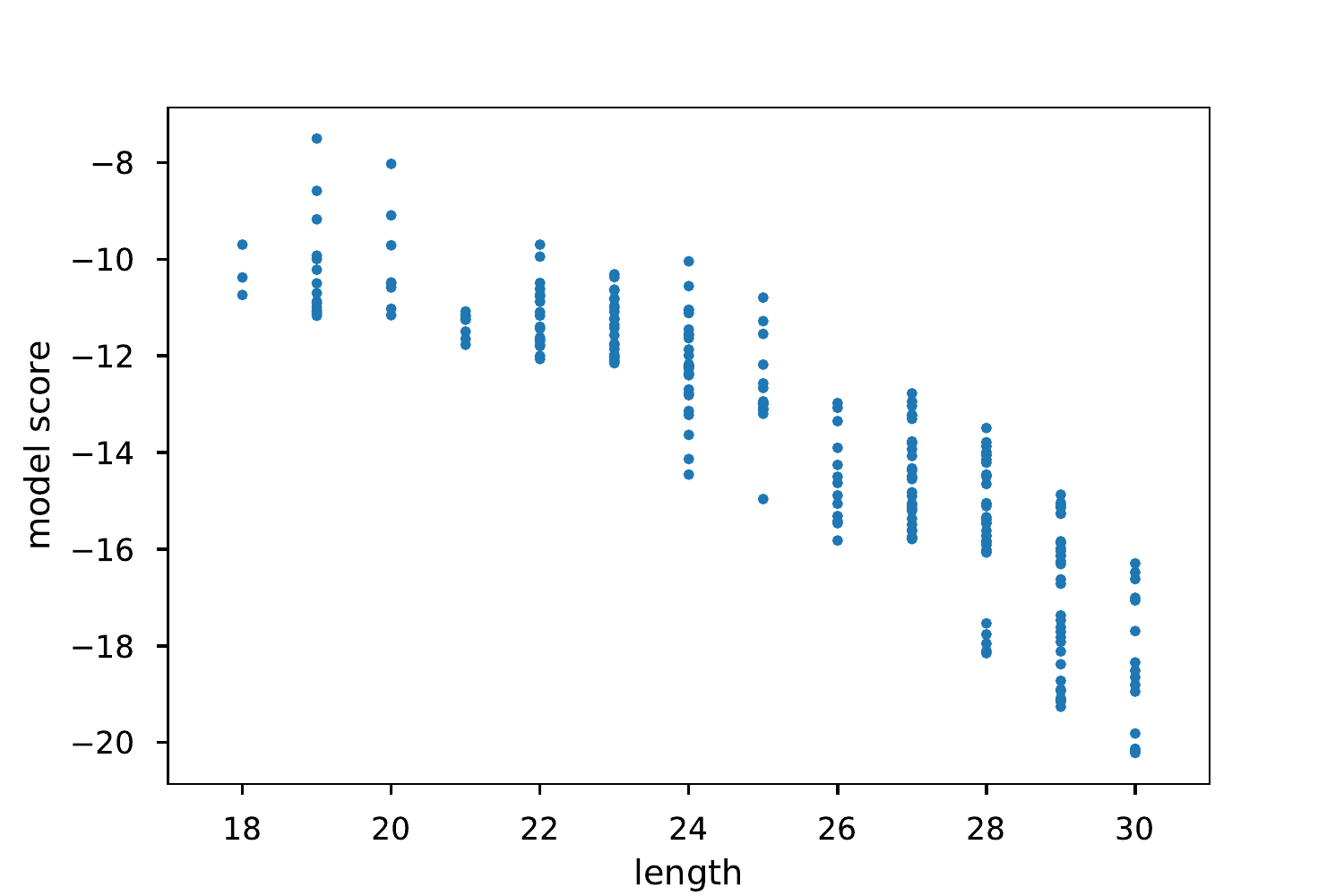}
  \vspace{-0.6cm}
\caption{Candidate lengths vs.~model score.
         This scatter plot is generated from 242 finished candidates when translated from
         one source sequence with beam size 80.}
\label{fig:length-model}
\end{figure}

The most popular translation quality metric, BLEU~\cite{BLEU:2002}, is defined as:
\begin{align}
\BLEU &= \bp \cdot \exp ( {1}/{4} \textstyle\sum_{n=1}^4 \log p_n )
\label{eq:bleu}\\
\text{where} \ \ \bp &= \min\{e^{1-1/\lratio},1\}
\label{eq:bp}\\
\text{where} \ \ \ \lratio &= {|\vecy|} / {|\vecy^*|}
\end{align}
Here $p_n$ are the $n$-gram precisions,
and $|\vecy|$ and $|\vecy^*|$ denote the hypothesis and  reference lengths,
while \bp is the \textit{brevity penalty} (penalizing short translations) and \lratio is the \textit{length ratio} \cite{shi+:2016,koehn+knowles:2017}, respectively.

With beam size increasing, $|\vecy|$ decrases,
which causes the length ratio to drop, as shown in Fig.~\ref{fig:curse-bleu}.
Then the brevity penalty term, as a function of the length ratio,
decreases even more severely.
Since $\bp$ is a key factor in BLEU, this explains why the beam search curse happens.\footnote{The length ratio is {\em not} just about BLEU:
  if the hypothesis length is only 75\% of reference length, something that should have been translated must be missing; i.e., bad adequacy.
  Indeed, \namecite{murray+chiang:2018} confirm the same phenomenon with METEOR.}

The reason why $|\vecy|$ decreases as beam size increases
is actually twofold:
\vspace{-0.1cm}
\begin{enumerate}
\item As beam size increases, the more candidates it could explore.
Therefore, it becomes easier for the search algorithm to find the \eos symbol.
Fig.~\ref{fig:average_length-beam} shows that
the \eos indices decrease steadily with larger beams.\footnote{Pre-neural SMT models, being probabilistic,
  also favor short translations (and derivations), 
  which is addressed by word (and phrase) reward. The crucial difference between SMT and NMT is
  that the former stops when covering the whole input, while the latter stops on emitting \eos.}
\vspace{-0.15cm}
\item Then, as shown in Fig.~\ref{fig:length-model},
shorter candidates have clear advantages {\em w.r.t.}~model score.
\end{enumerate}
Hence, as beam size increases, the search algorithm will generate shorter candidates,
and then prefer even shorter ones among them.\footnote{\namecite{murray+chiang:2018}
  attribute the fact that beam search prefers shorter candidates to the {\em label bias problem} \cite{lafferty+:2001} due to NMT's local normalization.}





%% file: methods.tex


We first review existing methods to counter the length problem and then propose new ones to address their limitations.
In particular, we propose to predict the target length
from the source sentence,
in order to choose a hypothesis with proper length.
\subsection{Previous Rescoring Methods}
RNNSearch~\cite{bahdanau+:2014} first introduces the \textit{length normalization} method,
whose score is simply the average model score:
\begin{equation}
\hat{\score}_{\lengthnorm}(\vecx,\vecy) = {\score(\vecx,\vecy)}/{|\vecy|}
\label{eq:leng_norm}
\end{equation}
This is the most widely used rescoring method 
since it is hyperparameter-free.

GNMT~\cite{wu+:2016} incorporates length and coverage penalty into
the length normalization method, while also adding two hyperparameters to adjust
their influences. (please check out their paper for exact formulas).

Baidu NMT~\cite{he+:2016} borrows the \textit{Word Reward} method from pre-neural MT,
which gives a reward $r$ to every word generated,
where $r$ is a hyperparameter tuned on the dev set:
\begin{equation}
\hat{\score}_{\mathrm{WR}}(\vecx,\vecy) = \score(\vecx,\vecy) + r\cdot |\vecy|
\label{eq:wr}
\end{equation}

Based on the above,
\namecite{huang+:2017} propose a variant called \textit{Bounded Word-Reward}
which only rewards up to an ``optimal'' length.
This length is calculated using a fixed ``generation ratio'' $\gratio$,
which is the ratio between target and source sequence length,
namely the average number of target words generated for each source word.
It gives reward $r$ to each word
up to a bounded length $L(\vecx,\vecy)=\min\{ |\vecy|, \gratio \cdot |\vecx| \}$:
\begin{equation}
\hat{\score}_{\mathrm{BWR}}(\vecx,\vecy) = \score(\vecx,\vecy) + r\cdot L(\vecx,\vecy)
\label{eq:pwr}
\end{equation}

\subsection{Rescoring with Length Prediction}
To remove the fixed generation ratio $\gratio$ from Bounded Word-Reward,
we use a 2-layer MLP,
which takes the mean of source hidden states as input,
to predict the generation ratio $\gratio^*(\vecx)$.
Then we replace the fixed ratio $\gratio$ with it,
and get our predicted length $L_{\pred}(\vecx) = \gratio^*(\vecx) \cdot |\vecx|$.
\subsubsection{Bounded Word-Reward}
With predicted length, the new predicted bound and final score would be:
\begin{align}
L^*(\vecx,\vecy) &= \min\{ |\vecy|, L_{\pred}(\vecx) \}\\
\hat{\score}_{\mathrm{BWR}^*}(\vecx,\vecy) &= \score(\vecx,\vecy) + r\cdot L^*(\vecx,\vecy)
\end{align}
While the predicted length is more accurate,
there is still a hyperparameter $r$ (word reward), so we design two methods below to remove it.

\subsubsection{Bounded Adaptive-Reward}
We propose \textit{Bounded Adaptive-Reward} to automatically calculate proper rewards based on the current beam.
With beam size $b$, the reward for time step
$t$ is the average negative log-probability
of the words in the current beam.


\begin{equation}
r_t = - ({1}/{b}) \textstyle \sum_{i=1}^b \log p(\word_i)
\end{equation}
Its score is very similar to (\ref{eq:pwr}):
\begin{equation}
\hat{\score}_{\AdaR}(\vecx,\vecy) = \score(\vecx,\vecy) + \textstyle\sum_{t=1}^{L^*(\vecx,\vecy)} r_t
\label{eq:adar}
\end{equation}

\subsubsection{BP-Norm}
Inspired by the BLEU score definition, we propose \textit{BP-Norm} method as follows:
\begin{equation}
\hat{\score}_{\bp}(\vecx,\vecy) = \log \bp + {\score(\vecx,\vecy)}/{|\vecy|}
\label{eq:bp_norm}
\end{equation}
\bp is the same brevity penalty term as in (\ref{eq:bp}).
Here, we regard our predicted length as the reference length.
The beauty of this method appears when we drop the logarithmic symbol in (\ref{eq:bp_norm}):
\begin{align*}
\exp(\hat{\score}_{\bp}(\vecx,\vecy)) &\!=\!\bp \!\cdot\! \exp \!\bigg( \frac{1}{|\vecy|} \textstyle\sum_{i=1}^{|\vecy|} \log p(y_i|...)\!\bigg)\\
&\!=\! \bp \!\cdot\! \big( \textstyle\prod_{i=1}^{|\vecy|} p(y_i|...) \big)^{{1}/{|\vecy|}}
\end{align*}
which is in the same form of BLEU score (\ref{eq:bleu}).


%% file: stop.tex

Besides rescoring methods, the stopping criteria (when to stop beam search) is also  important,
for both efficiency and accuracy.
\subsection{Conventional Stopping Criteria}
By default, OpenNMT-py~\cite{klein+:2017} stops when the topmost beam candidate stops,
because there will not be any future candidates with higher model scores.
However, this is not the case for other rescoring methods;
e.g., the score of length normalization (\ref{eq:leng_norm}) could still increase.

Another popular stopping criteria, used by RNNSearch ~\cite{bahdanau+:2014},
stops the beam search when exactly $b$ finished candidates have been found.
Neither method is optimal.

\subsection{Optimal Stopping Criteria}
For Bounded Word-Reward,
\namecite{huang+:2017} introduces a provably-optimal stopping criterion that could stop both early and optimally.
We also introduce an optimal stopping criterion for BP-Norm.
Each time we generate a finished candidate,
we update our best score $\hat{\score}^\star$.
Then, for the topmost beam candidate of time step $t$, we have:
\begin{equation}
\hat{\score}_{\bp} = \frac{\score_{t,0}}{t} + \min \{ 1-\frac{L_{\pred}}{t},0 \} \leq \frac{\score_{t,0}}{R}
\label{eq:bp_optim}
\end{equation}
where $R$ is the maximum generation length.
Since $\score_{t,0}$ will drop after time step $t$,
if $\frac{\score_{t,0}}{R} \leq \hat{\score}^\star$, we reach optimality.
This stopping criterion could also be applied to length normalization (\ref{eq:leng_norm}).

Meawhile, for Bounded Adaptive-Reward, we can have a similar optimal stopping criterion:
If the score of topmost beam candidate at time step $t>L_{\pred}$ is lower than $\hat{\score}^\star$,
we reach optimality.
\begin{proof}
The first part of $\hat{\score}_{\AdaR}$ in (\ref{eq:adar}) will decrease after time step $t$,
while the second part stays the same when $t>L_{\pred}$.
So the score in the future will monotonically decrease.
\end{proof}

%% file: exps.tex

Our experiments are on Chinese-to-English translation task,
based on the OpenNMT-py codebase.\footnote{\url{https://github.com/OpenNMT/OpenNMT-py}}
We train our model on 2M sentences, and apply BPE~\cite{sennrich+:2015} on both sides,
which reduces Chinese and English vocabulary sizes down to 18k and 10k respectively.
We then exclude pairs with more than 50 source or target tokens.
We validate on NIST 06 and test on NIST 08 (newswire portions only for both).
We report case-insensitive, 4 reference BLEU scores. 

We use 2-layers bidirectional LSTMs for the encoder.
We train the model for 15 epochs, and choose the one with lowest perplexity on the dev set.
Batch size is 64; both word embedding and hidden state sizes 500; and dropout 0.3.
The total parameter size is 28.5M.



\subsection{Parameter Tuning and Results}

\begin{figure}[t]
\centering
\vspace{-0.6cm}
\includegraphics[width=0.50\textwidth]{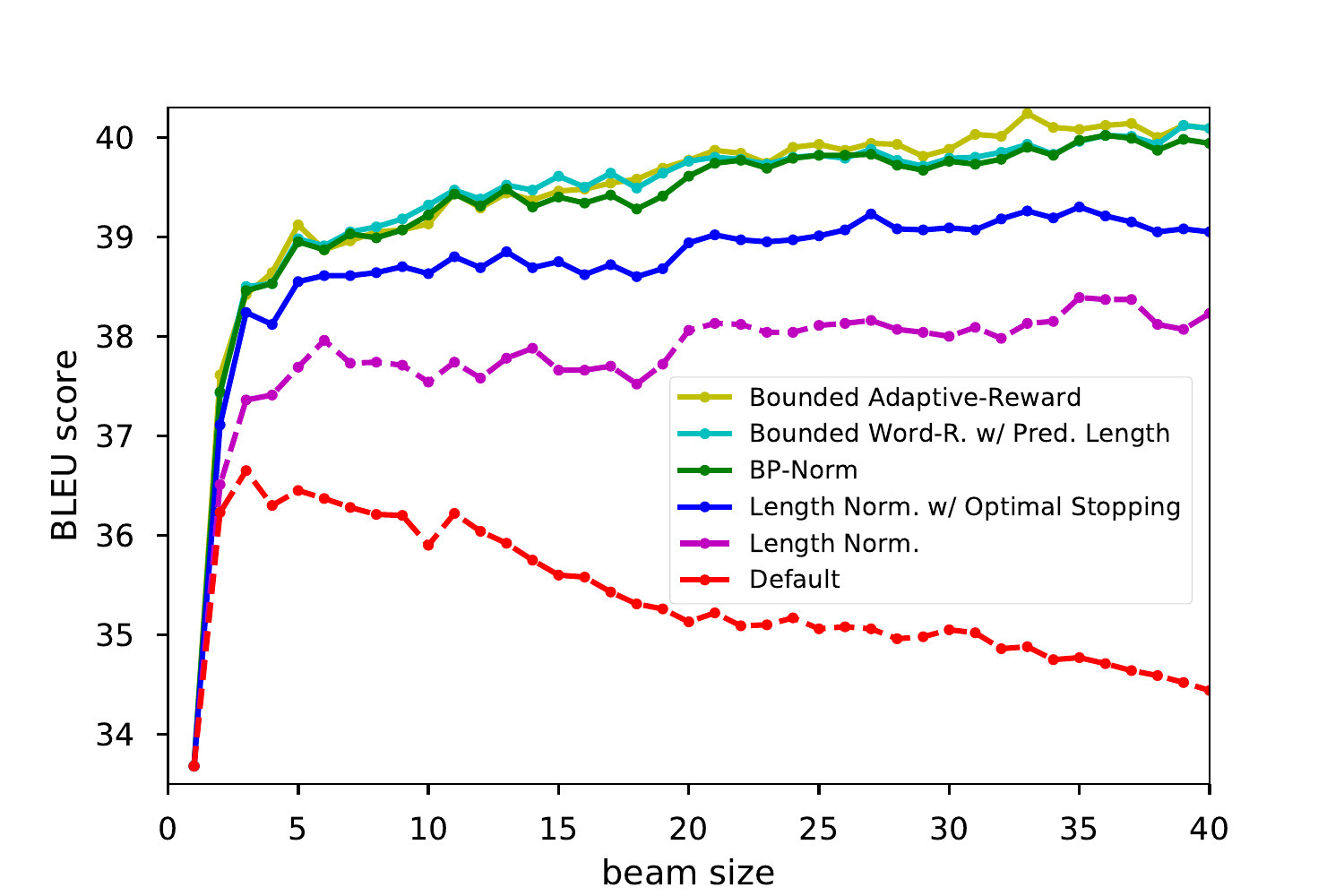}\\[-0.1cm]

\includegraphics[width=0.50\textwidth]{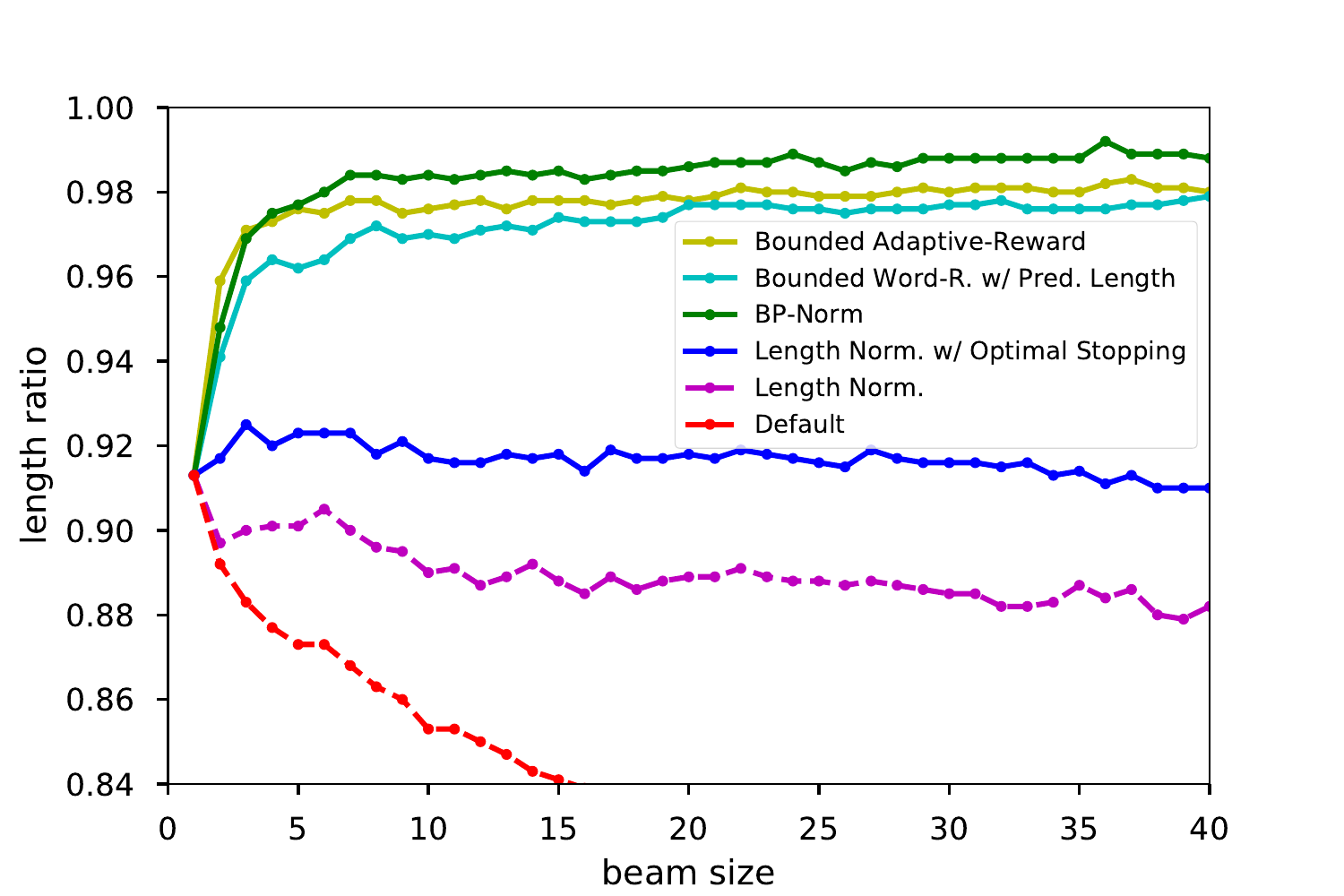}
\vspace{-0.6cm}
\caption{
The BLEU scores and length ratios ($\lratio=|\vecy|/|\vecy^*|$) of various rescoring methods.
}
\label{fig:results-length}
\end{figure}

\begin{figure}[t]
\centering
\vspace{-0.6cm}
\includegraphics[width=0.50\textwidth]{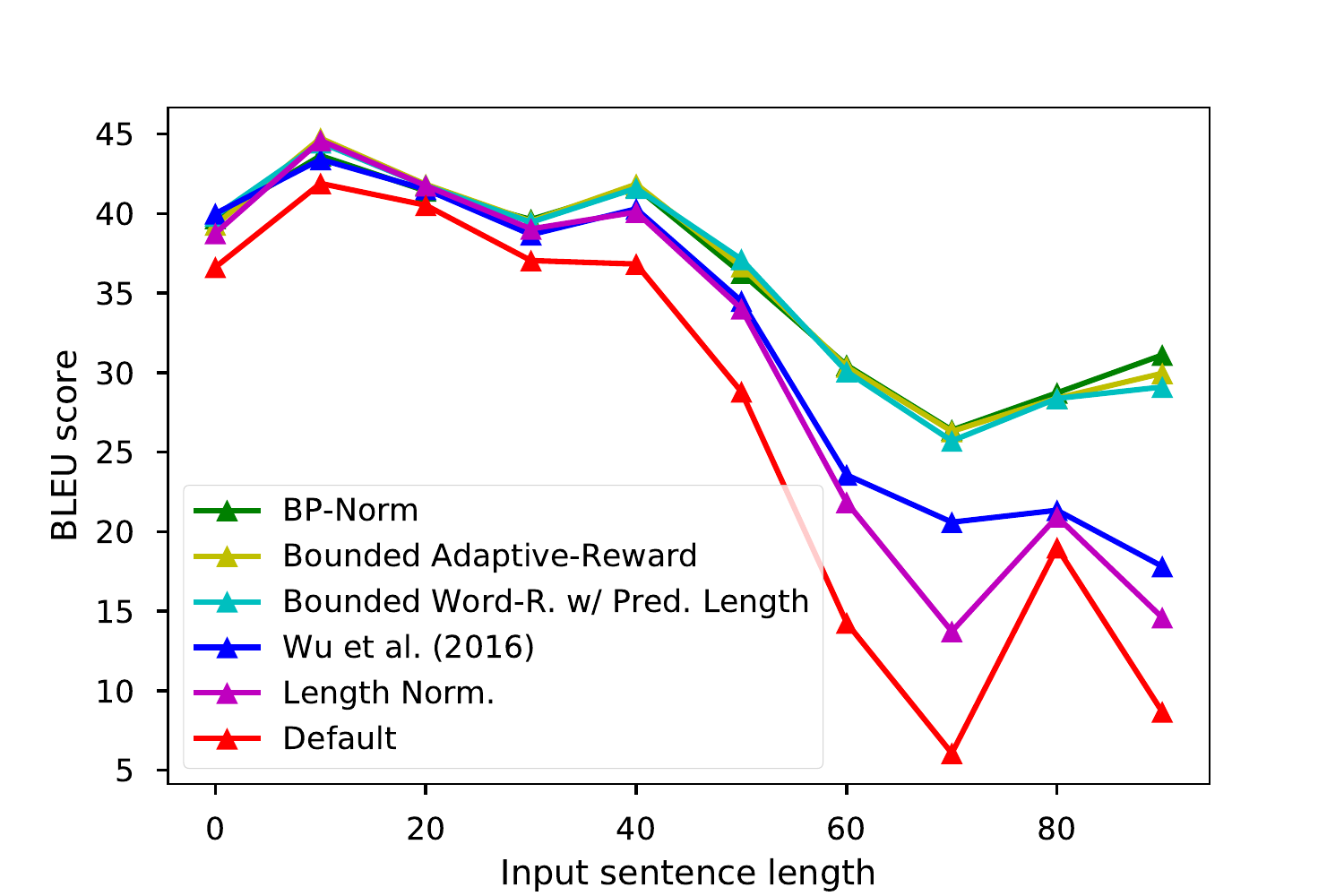}\\[-0.07cm]
\includegraphics[width=0.50\textwidth]{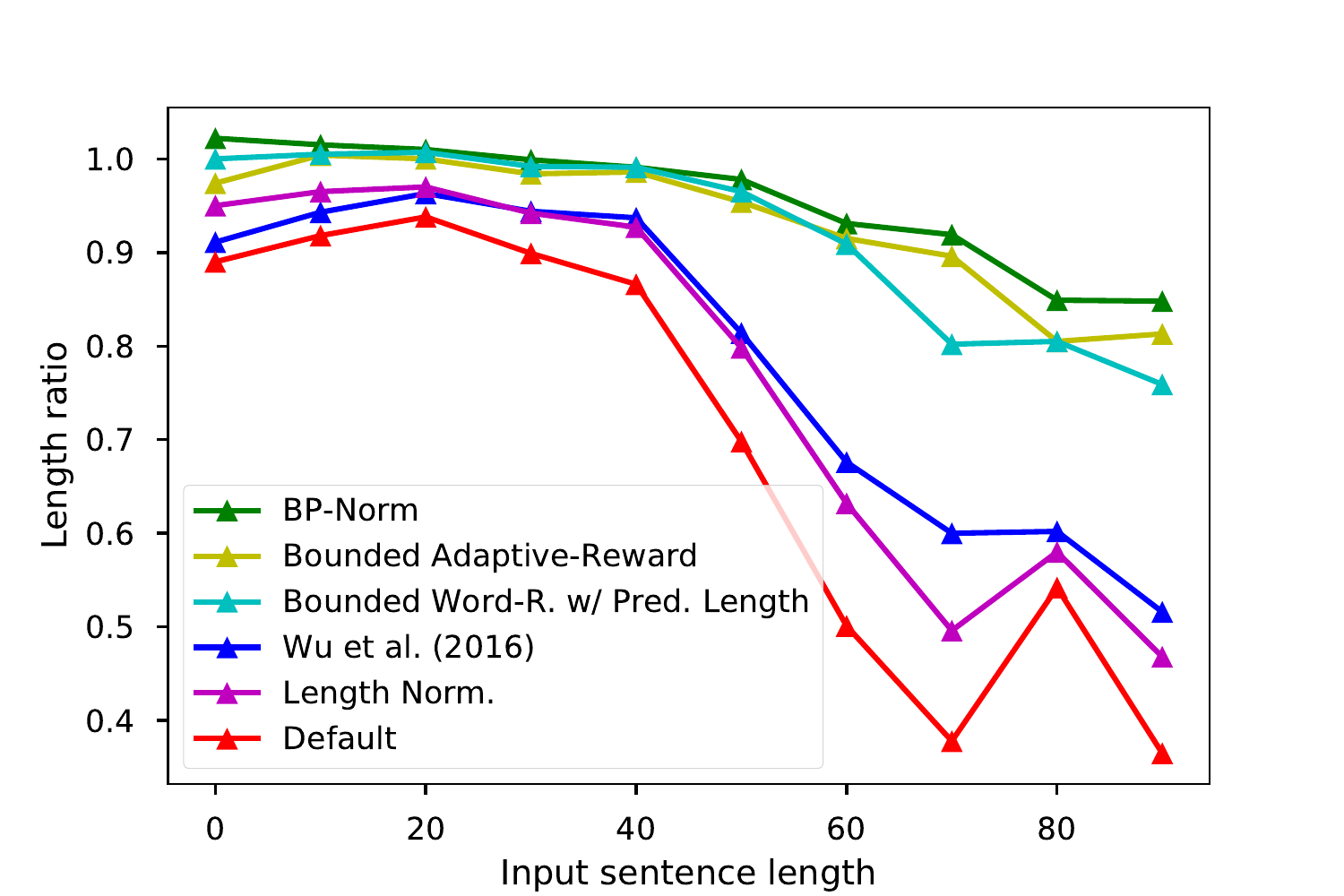}
\vspace{-0.6cm}
\caption{
BLEU scores and length ratios on the dev set over various input sentence lengths.
}
\label{fig:leng-ratio}
\end{figure}



\begin{table}
\resizebox{0.99\columnwidth}{!}{
\begin{tabular}{lcccc}
\toprule
 \multirow{2}{*}{Small beam ($b=14,15,16$)} & \multicolumn{2}{c}{dev} & \multicolumn{2}{c}{test}\\
 \cmidrule(lr){2-3}\cmidrule(lr){4-5}
 & BLEU & ratio & BLEU & ratio \\
\hline
Moses ($b$=70)& 30.14 & - & 29.41 & - \\
Default ($b$=5) & 36.45 & 0.87 & 32.88 & 0.87 \\
\hline
Length Norm. & 37.73 & 0.89 & 34.07 & 0.89 \\
+ optimal stopping$^*$ & 38.69 & 0.92 & 35.00 & 0.92 \\
\namecite{wu+:2016} $\alpha$=$\beta$=0.3 & 38.12 & 0.89 & 34.26 & 0.89 \\
Bounded word-r. $r$=1.3 & 39.22 & 0.98 & 35.76 & 0.98 \\
\hline
\textbf{with predicted length} &&&&\\
\hline
Bounded word-r. $r$=1.4$^*$ & \textbf{39.53} & 0.97 & 35.81 & 0.97 \\
Bounded adaptive-reward$^*$ & 39.44 & 0.98 & 35.75 & 0.98 \\
BP-Norm$^*$ & 39.35 & 0.98 & \textbf{35.84} & 0.99 \\
\bottomrule
\end{tabular}
}
\\[0.2cm]
\resizebox{0.99\columnwidth}{!}{
\begin{tabular}{lcccc}
\toprule
 \multirow{2}{*}{Large beam ($b=39,40,41$)} & \multicolumn{2}{c}{dev} & \multicolumn{2}{c}{test}\\
 \cmidrule(lr){2-3}\cmidrule(lr){4-5}
 & BLEU & ratio & BLEU & ratio \\
\hline
Moses ($b$=70)& 30.14 & - & 29.41 & - \\
Default ($b$=5) & 36.45 & 0.87 & 32.88 & 0.87 \\
\hline
Length Norm. & 38.15 & 0.88 & 34.26 & 0.88 \\
+ optimal stopping$^*$ & 39.07 & 0.91 & 35.14 & 0.91 \\
\namecite{wu+:2016} $\alpha$=$\beta$=0.3 & 38.40 & 0.89 & 34.41 & 0.88 \\
Bounded word-r. $r$=1.3 & 39.60 & 0.98 & 35.98 & 0.98 \\
\hline
\textbf{with predicted length} &&&&\\
\hline
Bounded word-r. $r$=1.4$^*$ & 40.11 & 0.98 & 36.13 & 0.97 \\
Bounded adaptive-reward$^*$ & \textbf{40.14} & 0.98 & \textbf{36.23} & 0.98 \\
BP-Norm$^*$ & 39.97 & 0.99 & \textbf{36.22} & 0.99 \\
\bottomrule
\end{tabular}
}
\caption{Average BLEU scores and length ratios over small and large beams. 
$\star$ indicates our methods.}
\label{tab:results}
\end{table}




We compare all rescoring methods mentioned above.
For the length normalization method,
we also show its results with \textit{optimal stopping}.

For Bounded Word-Reward method with and without our predicted length,
we choose the best $r$ on the dev set seperately.
The length normalization used by \namecite{wu+:2016} has two hyper-parameters,
namely $\alpha$ for length penalty and $\beta$ for coverage penalty.
We jointly tune them on the dev set, and choose the best config. 
($\alpha$=0.3,\ $\beta$=0.3).

Figure \ref{fig:results-length} show our results on the dev set.
We see that our proposed methods get the best performance on the dev set,
and continue growing as beam size increases.
We also observe that optimal stopping boosts the performance of length normalization method by around +0.9 BLEU.
In our experiments, we regard our predicted length as the
\textit{maximum generation length} in (\ref{eq:bp_optim}).
We further observe from Fig.~\ref{fig:leng-ratio}
that our methods keep the length ratio close to 1,
and greatly improve the quality on longer input sentences,
which are notoriously hard for NMT \cite{shen+:2016}.

Table~\ref{tab:results} collects our results on both dev and test sets.
Without loss of generality, we show results with both small and large beam sizes,
which average over $b$=14,15,16 and $b$=39,40,41, respectively.

\subsection{Discussion}

From Table~\ref{tab:results}, we could observe that
 with our length prediction model,
Bounded word-reward method gains consistent improvement.
On the other hand, results from length normalization method show that
optimal stopping technique gains significant improvement by around +0.9 BLEU.
While with both, our proposed methods beat all previous methods,
and gain improvement over hyperparameter-free baseline (i.e. length normalization)
by +2.0 BLEU.

Among our proposed methods,
Bounded word-reward has the reward $r$ as an hyper-parameter,
while the other two methods get rid of that.
Among them, we recommend the BP-Norm method,
because it is the simplest method, and yet works equally well with others.